\documentclass{article}

\usepackage[preprint]{neurips_2024}
\usepackage{algorithm}

\usepackage{amsmath, amssymb}

\usepackage[most]{tcolorbox}

\newcommand{\nablatilde}{\tilde{\nabla}}

\newcommand{\nablafwd}{\tilde{\nabla}^{+}}

\newcommand{\nablacentral}{\tilde{\nabla}^{\pm}}

\usepackage[noend]{algcompatible}

\title{Scaling Recurrent Neural Networks to a Billion Parameters with Zero-Order Optimization}

\usepackage[utf8]{inputenc} 
\usepackage[T1]{fontenc}    
\usepackage{hyperref}       
\usepackage{url}            
\usepackage{booktabs}       
\usepackage{amsfonts}       
\usepackage{nicefrac}       
\usepackage{microtype}      
\usepackage{xcolor}         
\usepackage{graphicx}
\usepackage{subcaption}  
\usepackage[capitalise]{cleveref}

\usepackage{pgfplots}
\pgfplotsset{compat=1.17}
\usepackage{pgfplotstable}

\author{%
  Francois Chaubard  \\
  Department of Computer Science\\
  Stanford University\\
  Stanford, CA 94305 \\
  \texttt{fchaubar@stanford.edu} \\
  \And
  Mykel J. Kochenderfer \\
  Department of Aeronautics and Astronautics\\
  Stanford University\\
  Stanford, CA 94305 \\
  \texttt{mykel@stanford.edu} \\
}

\begin{document}

\maketitle

\begin{abstract}
  During inference, Recurrent Neural Networks (RNNs) scale constant in both FLOPs and GPU memory with increasing context length, as they compress all prior tokens into a fixed-size memory. In contrast, transformers scale linearly in FLOPs and, at best, linearly in memory during generation, since they must attend to all previous tokens explicitly. Despite this inference-time advantage, training large RNNs on long contexts remains impractical because standard optimization methods depend on Backpropagation Through Time (BPTT). BPTT requires retention of all intermediate activations during the forward pass, causing memory usage to scale linearly with both context length and model size. In this paper, we show that Zero-Order Optimization (ZOO) methods such as Random-vector Gradient Estimation (RGE) can successfully replace BPTT to train RNNs with convergence rates that match, or exceed BPTT by up to 19 fold, while using orders of magnitude less memory and cost, as the model remains in inference mode throughout training. We further demonstrate that Central-Difference RGE (CD-RGE) corresponds to optimizing a smoothed surrogate loss, inherently regularizing training and improving generalization. Our method matches or outperforms BPTT across three settings: (1) overfitting, (2) transduction, and (3) language modeling. Across all tasks, with sufficient perturbations, our models generalize as well as or better than those trained with BPTT, often in fewer steps. Despite the need for more forward passes per step, we can surpass BPTT wall-clock time per step using recent advancements such as FlashRNN and distributed inference.
\end{abstract} 

\section{Introduction}
While transformers \citep{vaswani2017attention} have dominated large-scale language modeling \citep{brown2020language, touvron2023llama, zhang2022opt} over the past half-decade, their compute and memory requirements during inference prompt a search for alternatives for the future. During both training and inference, transformers compute pairwise attention across all $C$ previous tokens, resulting in quadratic complexity \( \mathcal{O}(C^2) \) in both compute and memory \citep{tay2020efficient, dao2022flashattention}. For instance, a ``small'' 1B-parameter Transformer with batch size $B = 2048$, context length $C = 1\text{M}$, and output size $O = 50{,}000$ would require 3.3 Petabytes of VRAM (or 44k H100s) assuming dense attention. If using linear attention mechanisms \citep{katharopoulos2020transformers, ahn2023linear, han2024demystify} such as Flash Attention \citep{dao2022flashattention}, this would still require 402TB of VRAM (or 5k H100s). 

Recurrent Neural Networks (RNNs), on the other hand, compress all necessary information of the past into a hidden state that is updated each time step instead of attending to all previous tokens per next token prediction. While this means the time to first token will scale linearly with the size of the prompt vs. transformers which would be constant, the time for all following tokens would be the same or faster as RNNs scale constant in FLOPs and GPU memory vs. linear and quadratic for transformers, respectively. Therefore, an equivalent 1B-parameter RNN would require only 28GB of VRAM which can fit easily on 1 single H100, or even smaller consumer GPUs, \emph{a savings of 14,000 times in cost per generated token}. While this is a huge savings, it requires a method that can efficiently train large RNNs with similar capacity as modern transformers, at batch sizes and sequence lengths at or greater than 1024, as is typically required to stabilize training and provide a sufficient context window for modern transformers to predict the next token accurately.


Scaling RNNs in model size, batch size, and context length using Backpropagation Through Time (BPTT) introduces a prohibitive memory bottleneck. BPTT requires storing all intermediate activations per time step for gradient computation, resulting in exploding VRAM complexity \( \mathcal{O}(BCA) \), where $B$ denotes the batch size, and $A$ denotes the sum of all activations per layer for one time step as well as vanishing and exploding gradients which dilute the gradient estimate. Early efforts to train without BPTT began with \citet{maeda2005simultaneous}, who applied Simultaneous Perturbation Stochastic Approximation (SPSA) \citep{spall1992multivariate} using Rademacher probes to train tiny RNNs (<32 hidden units) in hardware. In 2007, \citet{xia2007research} used Particle Swarm Optimization (PSO) to train small Elman networks for load forecasting. In 2016, \citet{rawal2016evolving} evolved compact LSTM-based memory architectures using an information-theoretic objective. More recently, Gradient approximation methods such as UORO~\citep{tallec2017unbiased}, DNI~\citep{jaderberg2017decoupled}, and e-prop~\citep{bellec2020solution} attempted to train small RNNs without unrolling.  \citet{vicol2021unbiased} introduced Persistent Evolution Strategies (PES) to provide unbiased estimates for long unrolls. Also, Koopman-inspired random feature networks~\citep{bolager2024gradient} and weight-biased perturbation methods~\citep{fernandez2024gradient} have achieved competitive results on small-scale RNNs with Zero-Order optimization. 


While no work has successfully scaled RNNs to billions of parameters and long context lengths, recent advances in zero-order training for large transformers have scaled to billions of parameters giving us inspiration to try these Zero-Order methods on large RNNs, such as MeZO~\citep{malladi2022mezo}, LeZO~\citep{lezo}, and SparseMeZO~\citep{malladi2023sparsemezo}. The major drawback of ZOO (and specifically RGE) is the reliance on a large number of forward passes to reduce perturbation noise to provide a sufficiently accurate gradient estimate. This wall-clock time can be intolerable if done sequentially and with legacy implementations. However, we sufficiently reduce these concerns with recent advancements in distributed training (e.g. Pytorch Distributed) and optimized CUDA kernels (specifically FlashRNN \citep{flashrnn}). First, we distribute the forward passes across a cluster of GPUs to achieve equal or better wall-clock time per step compared to BPTT. Then our bottleneck on wall-clock time per step is the time it takes for a single forward pass, which can still be quite slow for long sequence lengths. Recently, FlashRNN introduced a fused CUDA and Triton implementation of traditional RNNs (LSTM, GRU, sLSTM) that optimizes memory access by caching weights in registers and shared memory, achieving up to \emph{50× speedup over vanilla PyTorch}. FlashRNN also supports 40× larger hidden sizes which can enable us to scale. It is worth noting that during inference, we may be able to further reduce latency by precomputing the final hidden state of common prompts (e.g. system prompts) to avoid this repeated computation and high-latency.

In this paper, we introduce a new framework on how to scale large RNNs by merging recent advancements in distributed optimization (e.g. Pytorch Distributed), fused kernel RNNs (e.g. FlashRNN), and Zero-Order Optimization to scale RNNs well to and beyond the billion parameter mark. Specifically, we investigate Random-Vector Gradient Estimation (RGE) as it has been shown that RGE can approximate the true gradient with sufficient perturbations  \citep{spall1992multivariate, kiefer1952stochastic, duchi2015optimal, nesterov2017random}. In RGE, given model weights \( \Theta \in \mathbb{R}^{|\Theta|} \) and loss function \(L\), the true gradient \( \nabla L(\Theta) \) is approximated using the average of directional derivatives, calculated by perturbing the model by random probes \( p \) sampled from a distribution \( \mathcal{D}^{|\Theta|} \). To parallelize forward passes, we implement a distributed RGE where the clean model \( \Theta \) is broadcast to multiple ranks, each assigned a seed to generate local Rademacher probes. Each worker computes their perturbed forward passes (plus and minus), and returns only a 2 scalar losses. The parameter server reconstructs the probes, estimates directional derivatives, and applies updates via Stochastic Gradient-Free Descent. This design avoids inter-rank upload of full gradient information, reducing VRAM to \( \mathcal{O}(Ba_{\max} + |\Theta| + ChunkSize) \), where chunk size is the size of communication between ranks and can be tuned for optimal training speeds. Coupling distributed RGE and FlashRNN, we are able to match or exceed the wall-clock time per step of BPTT and Transformers with far simpler implementation to achieve scale. 

We apply this solver to scale to billion parameter RNNs in 3 domains. First, we compare the properties of CD-RGE compared to BPTT in a non-stochastic setting, by overfitting a DNC on a single batch and compare the number of steps to achieve near 0 loss. DNCs are notoriously hard to train as they have a complex memory matrix interaction that can be difficult for optimizers to solve which makes overfitting an interesting non-convex problem for the solver to navigate. As shown in \Cref{fig:bptt_vs_fda}, we find that with sufficient perturbations per step (e.g. 512), CD-RGE can achieve near 0 loss in 19x fewer steps. And we see monotonically better convergence with more perturbations per step. Second, we train LSTMs and DNCs on stochastic transduction tasks with narrow generalization (e.g. COPY, REVERSE, and ROLLING SUM) where we train with context lengths 2 to 10 randomly, and validate generalization to context lengths between 11 to 60. In all cases and model sizes, CD-RGE matched or outperformed BPTT. Finally, we extend this to a stochastic, broad generalization task, language modeling (e.g. Penn Tree Bank), where we train LSTMs and observe again that CD-RGE trained models, match or outperform BPTT trained models.

The remainder of the paper proceeds as follows: we provide background on Zero-Order Optimization; detail our algorithm and distributed training framework; present empirical results on our three sets of tasks, comparing convergence rates, VRAM usage, and inference cost relative to BPTT and Transformer baselines; provide intuition as to why RGE outperforms at specific $\varepsilon$; prove that optimizing the finite difference is map reducible to optimizing a smoothed surrogate and therefore is an implicit regularizer; and finally, discuss broader implications for meaningfully reduced FLOPs and VRAM (and thus cost) per generated token.

Our contributions are summarized as follows:
\begin{itemize}
  \item We show that Central-Difference RGE is able to efficiently train billion parameter RNNs (LSTMs and DNCs) on a single small GPU (A40 with 46GB of VRAM).
  \item We show that Central-Difference RGE can exceed the convergence rates of BPTT by up to 19 fold, while requiring orders of magnitude less VRAM and cost compared to BPTT and similar-sized transformer.
  \item We show that CD-RGE can distribute workloads horizontally, achieving wall-clock times per step similar to BPTT.
  \item We open-source our code at \href{https://github.com/Fchaubard/zero_order_rnn}{https://github.com/Fchaubard/zero\_order\_rnn} for reproducibility and further research.
\end{itemize}

\section{Zero-Order Optimization}
The earliest approaches to ZOO were based on finite difference methods, dating back to the 1950s with the work of \citet{kiefer1952stochastic}. These techniques estimate gradients by querying the objective function multiple times while perturbing the model differently each time. Two primary variants exist; forward-difference RGE (FD-RGE) \citep{duchi2015optimal} and and central-difference RGE (CD-RGE) \citep{nesterov2017random}, each with their own tradeoffs. For a give $\theta \in $

FD-RGE's directional derivative estimate is given by:
\begin{equation}
\nablafwd L(\Theta) = \frac{L(\Theta + \epsilon p) - L(\Theta)}{\epsilon} p,
\label{eq:fde}
\end{equation}
where \( \epsilon \) is a small positive scalar. 

CD-RGE's directional derivative estimate is given by:
\begin{equation}
\nablacentral L(\Theta) = \frac{L(\Theta + \epsilon p) - L(\Theta - \epsilon p)}{2\epsilon} p,
\end{equation}
and is unbiased under symmetric probe distributions.

Others \citep{spall1992multivariate, duchi2015optimal, nesterov2017random} have shown that by averaging \(n_{\text{pert}}\) directional derivative estimates
obtained from i.i.d.\ random probes \(p \sim \mathcal{D}^{|\Theta|}\),
we obtain a zeroth-order estimator \(\nablatilde L(\Theta)\)
whose mean approaches the true gradient \(\nabla L(\Theta)\)
as \(\epsilon \to 0\), and whose estimation variance
scales as \(\mathcal{O}(1/n_{\text{pert}})\) under mild moment conditions. Specifically, assuming that the loss \(L:\mathbb{R}^{|\Theta|}\!\to\!\mathbb{R}\)  
is three–times continuously differentiable with an \(L\!\)-Lipschitz gradient, and \(p\sim\mathcal{D}^{|\Theta|}\) be an i.i.d. probe with  
\(\mathbb{E}[p]=0\) and \(\mathbb{E}[pp^{\!\top}]=I\) then the following estimator properties hold for FD-RGE and CD-RGE given $\epsilon$ perturbation size and $n_{pert}$ perturbations:

      
\textbf{Bias of FD-RGE.}  
\begin{equation}
  \mathbb{E}\!\bigl[\nablafwd L(\Theta)\bigr]
  \;=\;
  \nabla L(\Theta)\;+\;\mathcal{O}(\epsilon).
  \label{eq:fd_bias}
\end{equation}

\textbf{Variance of FD-RGE.}  
\begin{equation}
  \mathbb{E}\!\Bigl[
     \bigl\|\nablafwd L(\Theta)-\nabla L(\Theta)\bigr\|_2^{2}
  \Bigr]
  \;=\;
  \mathcal{O}(\epsilon^{2})
  \;+\;
  \mathcal{O}\!\bigl(1/n_{\text{pert}}\bigr)
  \label{eq:fd_variance}
\end{equation}

\textbf{Bias of CD-RGE.}  
\begin{equation}
  \mathbb{E}[\nablacentral L(\Theta)]
    \;=\;
    \nabla L(\Theta)
    +\mathcal{O}(\epsilon^{2}).
  \label{eq:cd_bias}
\end{equation}

\textbf{Variance of CD-RGE.}  
\begin{equation}
  \mathbb{E}\!\Bigl[
     \bigl\|\nablacentral L(\Theta)-\nabla L(\Theta)\bigr\|_2^{2}
  \Bigr]
  \;=\;
  \mathcal{O}(\epsilon^{4})
  \;+\;
  \mathcal{O}\!\bigl(1/n_{\text{pert}}\bigr).
  \label{eq:cd_variance}
\end{equation}

\noindent
Thus, by increasing $n_{pert}$ and decreasing $\varepsilon$ we can reduce noise and more closely approximate the true gradient. We note FD-RGE is ``cheaper'' because we can reuse the clean query for all perturbations giving us the relation $n_{pert} = n_{queries} - 1$. However, it incurs a larger bias term; \(\mathcal{O}(\epsilon)\). To achieve the same $n_{pert}$, CD-RGE requires double the amount of queries in the limit compared to FD-RGE. However, it suppresses the bias to \(\mathcal{O}(\epsilon^{2})\) as the second order terms in the Taylor expansion cancel. Given a fixed query budget, we observe CD-RGE outperforms FD-RGE meaningfully. 




For choice of distribution $D$ for our probe $p_i$, any zero-mean, isotropic noise will maintain the convergence properties above (\eqref{eq:fd_bias} to \eqref{eq:cd_variance}). We evaluate three such distributions on the overfitting studies: Uniform ($U \sim [-1,1]$), Normal ($N(0,1)$), and Rademacher probes ($ p_i \in \{-1, +1\}^{|\Theta|} $). We observe Normal and Rademacher to be roughly equivalent in convergence rate, and Uniform slightly underperforms. Rademacher has been shown \citep{spall1992multivariate} to be optimal in high $|\Theta|$ dimensions and with large $n_{pert}$ perturbations. Rademacher vectors exhibit constant norm across samples, unlike Gaussian probes whose squared norms follow a chi-squared distribution, introducing additional variance in our estimate. Moreover, Rademacher probes can be stored as bit vectors, yielding a 32-fold memory saving relative to 32-bit floats. 

The relationship between our perturbation size $\varepsilon$ and step size $\eta$ is critical to achieving stable convergence. \Citet{spall1992multivariate} advises to relate them to a ratio of around $\eta/\varepsilon \approx 1e^{3}$ meaning we step 1,000 times further than we measured with our queries. While this may be appropriate in low-curvature loss functions, we find this to be very unstable in highly non-convex loss functions. This may be the nuance why so many studies determined RGE unstable when training large models from raw weights \citep{periyasamy2024guided,pcspsa, malladi2022mezo}. If we instead tie our $\eta$ and $\varepsilon$ to be the same, i.e. $\eta=\varepsilon$, all throughout training, and use the Rademacher distribution to concentrate our queries to the shell of the sphere, then we are essentially committing to taking a step of size $\eta=\varepsilon$ at each step, then measuring at that distance all around the sphere for dips in loss, then stepping at that distance the difference-weighted average of the probes. As long as $n_{pert}$ is sufficiently large, and $\varepsilon$ is small enough to fit inside the target basin, we find this procedure ``jumps'' much farther than a typical Stochastic Gradient Descent (SGD) step, with stability and confidence, even when gradients are unavailable or unstable in the interior of the sphere (e.g. with BPTT in long-sequence RNNs for example, or large quantization errors). 


An added benefit of setting $\eta=\varepsilon$ is that $\varepsilon$ and $\eta$ now cancel each other out to give us our final simple update equation which is more numerically stable:
\begin{equation}
\Theta' = \Theta - \frac{1}{2n_{\text{pert}}} \sum_{i=1}^{n_{\text{pert}}} ( L(\Theta + \epsilon p_i) - L(\Theta - \epsilon p_i) ) p_i,  
\label{eq:fdras_update_rule}
\end{equation}

Mathematically, we can interpret finite difference methods as optimizing a smoothed surrogate for the loss:
\begin{equation}
L_{\epsilon}(\Theta) := \mathbb{E}_{p \sim \mathcal{D}}[L(\Theta + \epsilon p)]
\label{eq:smoothed_surrogate_loss}
\end{equation}
where \( \mathcal{D} \) is a zero-mean probe distribution such as the Rademacher or standard Normal. This formulation corresponds to a convolution of the original loss \( L \) with the scaled probe distribution:
\begin{equation}
L_{\epsilon}(\Theta) = (L * \rho_{\epsilon})(\Theta),
\quad \text{where } \rho_{\epsilon}(x) := \frac{1}{\epsilon^d} \rho\left( \frac{x}{\epsilon} \right),
\label{eq:conv_surrogate_loss}
\end{equation}
and \( \rho \) is the probability density function of \( \mathcal{D} \). 
This convolution replaces sharp valleys/spikes with smoother transitions. A higher $\varepsilon$ results in more smoothing, higher bias, lower variance, while a lower $\varepsilon$ results in less smoothing, lower bias, and higher variance. 

Additionally, if \( p \sim \{-1, +1\}^d \) (Rademacher), then \( \|p\|_2 = \sqrt{d} \), and we are no longer exploring the interior of the ball, only the outer shell, and the expectation becomes:\[
L_\epsilon(\theta) \approx \mathbb{E}_{p \sim \mathbb{S}^{d-1}} \left[ L(\theta + \epsilon p) \right]
\] where \( \mathbb{S}^{d-1} \) denotes the unit sphere in \( \mathbb{R}^d \). Thus, our finite difference formulation can be seen as implicitly optimizing a smoothed surrogate loss which is the difference weighted average probe sampled at the shell of a hypersphere of radius \( \epsilon \sqrt{d} \).

\section{Method: Distributed CD-RGE}

As summarized in \cref{alg:fdras_algo}, we divide our compute over $w$ total ranks, with rank 0 designated as the parameter update rank, and ranks $\{1, \ldots, w\}$ as worker ranks. At $w=1$, this becomes the sequential (local) version of the algorithm. Our training framework applies central-difference random gradient estimation with Rademacher probes (CD-RGE) in a distributed manner where rank 0 sends the most up-to-date model to all other ranks and a seed per perturbation to be used by the worker to deterministically generate each workers random Rademacher probe. Then all ranks forward pass their models. Then rank 0 receives back all scalar loss values, regenerates each probe, scales it by the finite difference, and updates the model, without ever fully instantiating any of the probes fully, just chunks at a time. 
\begin{algorithm}[H]
\caption{DIST.CDRGE\_STEP$(\theta,\varepsilon,x,n,w)  \rightarrow \theta$\label{alg:fdras_algo}
}

\begin{algorithmic}[1] 
\STATE \textbf{Input:} parameters $\theta\in\mathbb{R}^d$, perturbation scale $\varepsilon>0$, input batch $x$,  $n$ perturbations, $w$ ranks
\STATE $\textit{r} \leftarrow \textsc{dist.get\_rank}()$
\STATE \textsc{dist.broadcast}$([\theta,x],\text{src}=0)$
\IF{$\textit{r}=0$}
        \STATE Create unique seeds $S=\{\,s_{r,m}\mid r=0,\dots,w;\;m=0,\dots,n/w\}$
        \STATE Create $L_{array}=\{\,[L_{-}=0,L_{+}=0]_{r,m}\mid r=0,\dots,w;\;m=0,\dots,n/w\}$
\ENDIF

\STATE \textsc{dist.scatter}$(S,\text{src}=0)$
\FORALL{$m = 1 \dots (n/w)$} 
    \STATE $\theta \leftarrow \textsc{ApplyProbe}(\theta,\;\epsilon,\;s_{r,m})$ 
    \STATE $L_{+} \leftarrow \mathcal{L}(\theta,x)$ 
    \STATE $\theta \leftarrow \textsc{ApplyProbe}(\theta,\;-2\epsilon,\;s_{r,m})$ 
    \STATE $L_{-} \leftarrow \mathcal{L}(\theta,x)$
    \STATE $\theta \leftarrow \textsc{ApplyProbe}(\theta,\;\epsilon,\;s_{r,m})$ 
    \STATE \textsc{dist.gather}$([L_{-}, L_{+}],\text{dst}=0)$ 
\ENDFOR

\IF{$\textit{r}=0$}    
    \STATE $L_{array} \leftarrow \textsc{flatten}(L_{array})$
    \FORALL{$i = 1 \dots n$} 
        \STATE $\alpha \leftarrow \dfrac{L_{+}^{(i)} - L_{-}^{(i)}}{2n}$ 
        \STATE $\theta \leftarrow \textsc{ApplyProbe}(\theta,\;\alpha,\;s_{i})$ 
    \ENDFOR
\ENDIF

\end{algorithmic}
\end{algorithm}

\vspace{-2.0em} 

\begin{algorithm}[H]
\caption{ $\textsc{APPLYPROBE}(\theta,\alpha,s) \rightarrow \theta$}
\begin{algorithmic}[1] 
\STATE \textsc{torch.manual\_seed}$(s)$
\FORALL{$\theta_i \in \theta$} 
    \STATE $\theta_i \leftarrow \theta_i + \alpha \cdot \bigl(2\!\cdot\!\textsc{Bernoulli}(0.5;\lvert\theta_i\rvert)-1\bigr)$
\ENDFOR
\end{algorithmic}
\end{algorithm}

This allows us to dedicate the entire GPU memory to only parameters and activations aside from a small constant sized chunk for buffer. This enables us to train our model at batch sizes, context lengths and parameter scales impracticable with BPTT on the same hardware.




\subsection{Training Memory and Compute}

Our Zero-Order framework shifts the bottleneck from memory to compute as memory is 18x more expensive. We compare our method to training RNNs with BPTT and training transformers with backpropagation below:
\vspace{-0.5em}
\setlength{\itemsep}{0pt}
\setlength{\parskip}{0pt}
\setlength{\parsep}{0pt}
\begin{itemize}
\vspace{0.3em}
    \item \textbf{Memory usage:} We shift from $\mathcal{O}(BCA)$ for BPTT and modern transformers (assuming fast-attention) to $\mathcal{O}(Ba_{\max})$, for $B$ batch size, $C$ context length, and $a_{\max}$, the largest activation across layers.
    \item \textbf{Compute cost:} We unfortunately shift from only one forward and backward pass to $2\cdot n_{\text{pert}}$ forward passes per step (e.g. 96 and 512). Probes are reconstructed $4\times$ but each is fast and parallelized on the GPU. Retention of the Rademacher probe would be quite small since it can be represented as bit vector if its preferred to exchange FLOPs and time for a small additional Memory and communication overhead.
    \item \textbf{Wall-clock time per step:} For both BPTT and CD-RGE, we must wait $\mathcal{O}(C\space n_{pert}/w)$ time for $w$ ranks to complete their forward passes, which is the biggest disadvantage RNNs have compared to transformers, which have $\mathcal{O}(1)$ wall-clock time per step as they parallelize across the sequence. 
\end{itemize}

While we increase the number of FLOPS by $\mathcal{O}(n_{\text{pert}})$ and wall-clock time by $\mathcal{O}(BC \space n_{\text{pert}}/r)$, we reduce the VRAM by $\mathcal{O}(C(A-a_{max}))$ which is substantial. We estimate the cost impact of this on our experiments in \Cref{tab:costs_rge_vs_bptt} and at much larger scales using theoretical values in \Cref{fig:vram_fdras_grid}.

\subsection{Inference Memory and Compute Characteristics}

RNNs in general have far superior inference memory and compute requirements. We compare RNNs to transformers below:
\begin{itemize}
    \item \textbf{Memory usage:} We shift from $\mathcal{O}(CA)$ for modern transformers (assuming fast-attention) to $\mathcal{O}(a_{\max})$, where $a_{\max}$ is the size of the largest activation per layer. 
    \item \textbf{Compute cost:} Compute shifts from $\mathcal{O}(C)$ FLOPs per generated token to $\mathcal{O}(1)$ which is a dramatic improvement.
    \item \textbf{Wall-clock time for first token:} The time to first token scales linearly for RNNs, while transformers scale constant time as transformers parallelize the compute for the first token. This is a huge benefit of transformers. However, we believe this can be ameliorated by caching off our system prompt hidden states to avoid recomputation. 
    \item \textbf{Wall-clock time for all other tokens:} RNNs will achieve the same or better performance compared to transformers as they require fewer flops and both are sequential operations after the first token.
\end{itemize}

\begin{table}[h]
\centering
\begin{tabular}{llccccc}
\toprule
\textbf{Method} & \textbf{Metric} & \textbf{100k} & \textbf{1M} & \textbf{10M} & \textbf{100M} & \textbf{1.1B} \\
\midrule
BPTT & VRAM (GB) & 0.8 & 4 & 45 & NaN & NaN \\
     & Time per step (s) & 0.1 & 0.17 & 33.7 & NaN & NaN \\
     & \$ per 100k step & \$0.04 & \$0.20 & \$2.34 & NaN & NaN \\
\midrule
CD-RGE@96 (seq) & VRAM (GB) & 0.5 & 1 & 4 & 25.3 & 39.5 \\
                & Time per step (s) & 5.7 & 6 & 7.8 & 24.3 & 145 \\
                & \$ per 100k steps & \$0.04 & \$0.07 & \$0.22 & \$1.33 & \$2.38 \\
\midrule
CD-RGE@512 (seq) & VRAM (GB) & 0.5 & 1 & 4 & 25.3 & 39.5 \\
                 & Time per step (s) & 30 & 31.1 & 40.7 & 132 & 777 \\
                 & \$ per 100k steps & \$0.11 & \$0.14 & \$0.31 & \$1.63 & \$4.13 \\
\midrule
CD-RGE@96 (dist@8) & VRAM (GB) & 4 & 8 & 32 & 202.4 & 316 \\
                   & Time per step (s) & 0.05 & 0.06 & 0.07 & 0.4 & 4.4 \\
                   & \$ per 100k steps & \$0.20 & \$0.40 & \$1.60 & \$10.12 & \$15.81 \\
\midrule
CD-RGE@512 (dist@8) & VRAM (GB) & 4 & 8 & 32 & 202.4 & 316 \\
                    & Time per step (s) & 0.25 & 0.29 & 0.29 & 1.87 & 19.3 \\
                    & \$ per 100k steps & \$0.20 & \$0.40 & \$1.60 & \$10.13 & \$15.85 \\
\bottomrule
\end{tabular}
\vspace{1em}
\caption{Actual VRAM usage (GB), time per step (s), and linearized cost estimate per 100k steps (USD) for different training strategies across increasing model sizes. All measurements assume batch size = 1024 and sequence length = 10. We compare CD-RGE at two different perturbation counts per step (96 and 512) and we compare if we do them sequentially on one GPU vs. distributed on a node with 8 GPUs (all on A40s for consistency).}
\label{tab:costs_rge_vs_bptt}
\end{table}

\subsection{Actual Training Memory, Compute, and Cost Comparisons}

\Cref{tab:costs_rge_vs_bptt} provides  VRAM usage and average wall-clock time per step at increasing model sizes comparing BPTT, sequential CD-RGE at 96 perturbations per step, and sequential CD-RGE at 512 perturbations where we do not distribute the work load and do all foreward passes on one A40 GPU. We also compare to Distributed CD-RGE where we parallelize the forward passes over a node with 8 A40 GPUs respectively. For all runs, we use sequence length of 10 and batch size of 1024. When a batch of 1024 does not fit on the GPU, we must split the batch into micro and macrobatches to fit the microbatch into memory then loop over the macrobatches accumulating gradient. This is the case for BPTT at model size 10M, where we must use a macrobatch of 128 to fit the batch onto one GPU and in the case for CD-RGE at 1.1B where we must split the batch across 4 macrobatches. This explains the sharp uptick in time in those cases. While macrobatches could distributed as well, we save the analysis of distributed BPTT and splitting of macrobatches with CD-RGE to future research. To estimate cost, we scrape current GPU prices on leading cloud providers and regress to get an estimated cost per GB-Hr of VRAM (which we estimate to be \$0.018) and an estimated cost per TFLOP-Hr (which we estimate to be \$0.001). We note that memory is 18x more expensive than compute over all GPU types.

\vspace{.4em}

First, we see that even on small sequence lengths (10) and a medium size model (10M parameters), we use 10x less memory by switching from BPTT to CD-RGE@96, 340x reduction in wall-clock time per step since the entire batch can easily fit in memory and we do not have to loop over macrobatches, and thus, we see a \emph{cost savings of 30x} per training step. And as demonstrated in most of our experiments, we see a similar or improved convergence rate with these settings. Second, we note that BPTT shoots up in VRAM, quickly saturating the GPU even on our medium size model (10M) and a very short sequence length requiring us to either use macrobatches or use gradient checkpointing both of which are very slow. Next, we see that even 1.1B models at any sequence length can fit into VRAM with RGE. Finally, we note that by distributing the compute over more ranks, we can achieve faster wall-clock times easily and actually beat BPTT's wall-clock time per step.

\section{Experiments}

\subsection{Overfitting to a single batch}
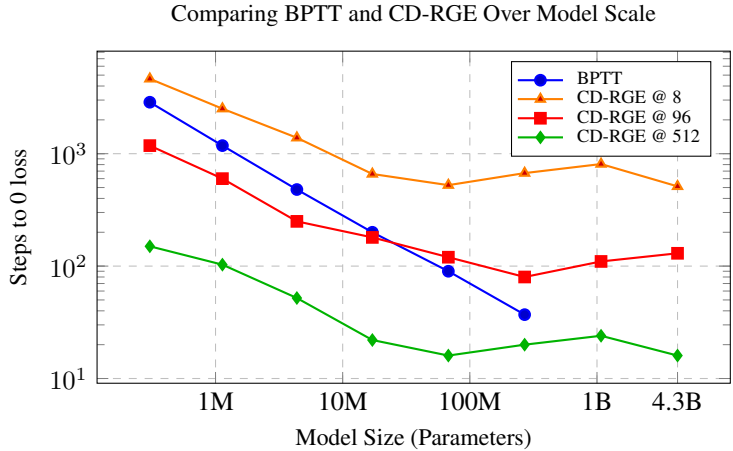
\begin{figure}[h]
    \centering
        \begin{tikzpicture}
        \begin{loglogaxis}[
            title={Comparing BPTT and CD-RGE Over Model Scale},
            xlabel={Model Size (Parameters)},
            ylabel={Steps to 0 loss},
            legend pos=north east,
            xmajorgrids=true,
            ymajorgrids=true,
            grid style=dashed,
            width=10cm,        
            height=6cm,        
            xtick={1e6,1e7,1e8,1e9,4.3e9},
            xticklabels={1M,10M,100M,1B,4.3B},
            legend style={
                font=\scriptsize,        
                row sep=-3pt,            
                column sep=3pt,          
                cells={anchor=west},     
                inner xsep=2pt,          
                inner ysep=2pt,          
                scale=0.8,               
            },
            legend cell align=left,
            title style={font=\small},      
            xlabel style={font=\small},     
            ylabel style={font=\small},
        ]
        
        \addplot+[mark=*, thick, blue] table {
        x y
        304357 2870
        1132901 1180
        4362853 480
        17114213 200
        67782757 90
        269783141 37
        };
        
        \addlegendentry{BPTT}
        
        \addplot+[mark=triangle*, thick, orange] table {
        x y
        304357 4635
        1132901 2514
        4362853 1386
        17114213 660
        67782757 525
        269783141 672
        1076437093 809
        4300357733 512
        };
        
        \addlegendentry{CD-RGE @ 8}
        \addplot+[
            mark=square*, 
            thick, 
            red, 
            mark options={fill=red}
        ] table {
        x y
        304357 1180
        1132901 600
        4362853 250
        17114213 180
        67782757 120
        269783141 80
        1076437093 110
        4300357733 130
        };
        
        \addlegendentry{CD-RGE @ 96}
        
        \addplot+[mark=diamond*, thick, green!70!black] table {
        x y
        304357 150
        1132901 103
        4362853 52
        17114213 22
        67782757 16
        269783141 20
        1076437093 24
        4300357733 16
        };
        
        \addlegendentry{CD-RGE @ 512}
        
        \end{loglogaxis}
        \end{tikzpicture}
    
    \caption{
        Iterations to overfit a fixed batch using BPTT versus CD-RGE on sequence length 100 at varying compute budgets. CD-RGE with 512 perturbations per step outperforms BPTT by up to 19× for small models (300k params) and 2× for larger models (270M params). BPTT cannot train models larger than 270M due to GPU memory constraints.
    }
    \label{fig:bptt_vs_fda}
\end{figure}

To compare convergence rates in a non-stochastic, non-convex setting, we compare CD-RGE to BPTT overfitting challenge. We select DNCs as they are notoriously difficult to train. We train DNCs ranging from tiny (300k) to very large (4.3B) and measure the number of steps required to achieve near 0 loss. As shown in \Cref{fig:bptt_vs_fda}, CD-RGE can meaningfully outperform BPTT provided enough perturbations. In the case of smaller models, this difference could be as large as 19 times faster with CD-RGE. We average performance over 5 runs but find that convergence between runs is within $\pm2$\%. 

\subsection{Transduction}

Now that we have shown CD-RGE can overfit to a single batch and achieve near 0 loss, we broaden to a stochastic, narrow training environment and a different model. We train LSTMs on COPY, REVERSE, and ROLLING SUM. In all cases, we provide a training context length from 1 to 10 and then measure how well the RNN can generalize to longer sequences (e.g. from 11 to 60). In \Cref{tab:lstm_results}, we include our results choosing the best performing model over a hyperparameter sweep for BPTT and CD-RGE and observe that CD-RGE is able to match or beat almost on all tasks. Additionally, often 96 perturbations is sufficient to match or even exceed BPTT performance while 512 beats about 70\% of the time. In the case of the medium size model on all tasks, BPTT quickly overfits, while CD-RGE seems to generalize better. Additionally, it is worth noting that BPTT is quite noisy while CD-RGE optimizes in a very smooth and stable way. We believe this is due to the tying of $\varepsilon$ and $\eta$. If we shrink $\varepsilon$ too small, we start to see similar instability and even divergence. 

\begin{table}[ht]
\centering

\label{tab:lstm_results}
\begin{tabular}{llccccc}
\toprule
\textbf{Task} & \textbf{Optimizer} & \textbf{100k} & \textbf{1M} & \textbf{10M} & \textbf{100M} \\
\midrule
Copy & BPTT          & 3.40 & 3.38 & 3.50 & -- \\
     & CD\textendash RGE@96  & 3.40 & 3.38 & 3.37 & 3.50 \\
     & CD\textendash RGE@512 & \textbf{3.37} & \textbf{3.35} & \textbf{3.36} & \textbf{3.40} \\
\midrule
Reverse & BPTT       & 3.87 & 3.40 & 3.48 & -- \\
        & CD\textendash RGE@96  & 3.35 & 3.35 & \textbf{3.36} & 3.50 \\
        & CD\textendash RGE@512 & \textbf{3.30} & \textbf{3.34} & 3.37 & \textbf{3.40} \\
\midrule
Add & BPTT           & \textbf{1.52} & \textbf{1.50} & 1.75 & -- \\
    & CD\textendash RGE@96   & 1.56 & 1.57 & 1.71 & 1.71 \\
    & CD\textendash RGE@512  & 1.56 & 1.56 & \textbf{1.70} & \textbf{1.70} \\
\midrule
PTB & BPTT           & 2.42 & \textbf{2.40} & 2.84 & -- \\
    & CD\textendash RGE@96   & 2.51 & 2.51 & \textbf{2.45} & 2.58 \\
    & CD\textendash RGE@512  & \textbf{2.37} & 2.50 & 2.50 & \textbf{2.47} \\

\bottomrule
\end{tabular}
\vspace{2em}
\caption{Validation loss (in nats) across our transduction tasks and language modeling task on 4 different model scales (from 100k parameters to 100M parameters) comparing BPTT to CD-RGE at 96 and 512 perturbations per step. Note, we can not fit the 100M parameter model on an A40 even with batch size 1 and gradient checkpointing, so can not compare. We see that often 96 perturbations is sufficient to match or even exceed BPTT performance, while in most cases CD-RGE with 512 perturbations outperforms.}
\end{table}

\subsection{Language Modeling}

\begin{figure}[ht]
    \centering
    \begin{tikzpicture}
    \begin{axis}[
        title={Validation Loss for BPTT and CD-RGE@512 Across Model Scales},
        xlabel={Training Step},
        ylabel={Validation Loss},
        legend pos=north east,
        xmajorgrids=true,
        ymajorgrids=true,
        grid style=dashed,
        width=10cm,
        height=6cm,
        xtick={100,200,300,400,500},
        ymin=2,
        ymax=5,
        legend style={
            font=\scriptsize,
            row sep=-3pt,
            column sep=3pt,
            cells={anchor=west},
            inner xsep=2pt,
            inner ysep=2pt,
            scale=0.8,
        },
        title style={font=\small},
        xlabel style={font=\small},
        ylabel style={font=\small},
    ]
\addplot+[mark=*, thick, blue] table {
50 4.46875
100 4.28125
150 4.125
200 4.0
250 3.84375
300 3.71875
350 3.625
400 3.5625
500 3.46875
};
\addlegendentry{CD-RGE (1M)}

\addplot+[
    mark=diamond*, 
    thick, 
    green!70!black, 
    mark options={fill=green!40!white}
] table {
50 3.28125
100 3.125
150 3.03125
200 2.8125
250 2.71875
300 2.984375
350 2.84375
400 2.671875
450 2.875
500 2.890625
};
\addlegendentry{BPTT (10M)}

\addplot+[
    mark=diamond*, 
    thick, 
    orange, 
    mark options={fill=orange}
] table {
50 3.96875
100 3.578125
150 3.390625
200 3.328125
250 3.214375
300 3.173125
350 3.100625
400 3.055253
450 3.028125
500 2.982625
};
\addlegendentry{CD-RGE (10M)}

\addplot+[mark=square*, thick, red] table {
50 3.203125
100 2.984375
150 2.8125
200 2.671875
250 2.609375
300 2.546875
350 2.53125
400 2.506875
450 2.4375
500 2.4175
};
\addlegendentry{CD-RGE (1.1B)}

\end{axis}
\end{tikzpicture}

\caption{Validation loss trajectories for training large LSTMs on Penn-Treebank dataset on next character prediction with BPTT and CD-RGE with 512 perturbations per store across model sizes 1M, 10M, and 1.1B. As you can see, RGE provides very smooth convergence, while BPTT overfit quickly and bounces unevenly even with Adam. Both BPTT and CD-RGE at 10M converge to similar loss values. However, larger model sizes converge faster and to lower loss values suggesting more capacity will benefit larger RNNs.}
\label{fig:ptb_val_loss_comparison_model_scale}
\end{figure}
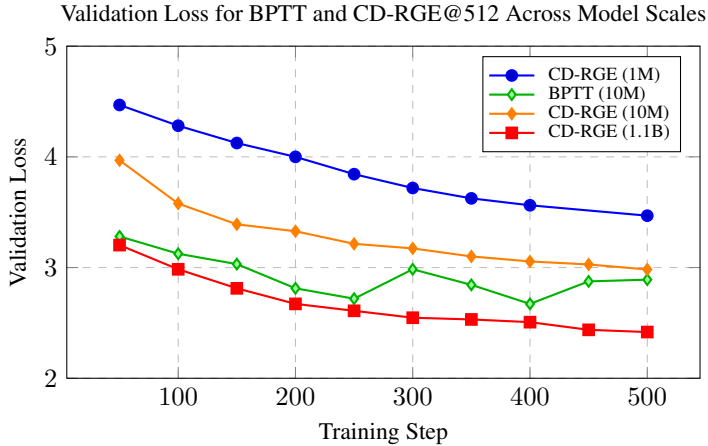

Finally, we broaden our experiments of CD-RGE to general language modeling. We do not aim to match state-of-the-art transformers in performance in this paper, instead only aim to show that BPTT and RGE perform similarly on the same task with the same model. We train LSTMs from 100k to 1.1B on the Penn-Treebank task \citep{penntreebank} to perform next token prediction, with batch size 1024 and sequence length 10 in all case. We measure that all of our model sizes are able match or exceed BPTT performance as shown in \Cref{tab:lstm_results}.  \Cref{fig:ptb_val_loss_comparison_model_scale} shows that we converge faster with larger model sizes, similar to the overfitting and transduction experiments. While Penn-Tree bank is a rather small dataset, so discerning generalization ability of larger RNNs is difficult and out of scope for this paper, we leave measuring of a more general scaling law for RNNs to future research. However, its clear that CD-RGE is able train larger RNNs whether a scaling law exists for LSTMs or not.  

\subsection{Implementation Details}

All models use a character-level tokenizer with an input embedding dimension of 32. For CD-RGE, we use Rademacher for all probe sampling. Although we experimented with Adam-based solvers for CD-RGE, performance degraded, likely due Adam inhibiting large shifts in direction that are often beneficial with larger steps. No experiments use learning rate schedules or curriculum learning. We always hold $\eta = \varepsilon$ throughout training for all runs. We explore $\varepsilon, \eta \in [0.1, 1\text{e}{-5}]$ and found $\varepsilon / \eta \in [1, 10]$ to be optimal. We sweep a learning rate from $10^{-5}$ to 0.1 for all models. As batch size and number of perturbations increase, we typically must increase $\eta$ to capture the benefit; conversely, larger models require smaller $\varepsilon$ all else held equal. We use a layer LSTM for all LSTM and DNC runs. For LSTM experiments, we use FlashRNN kernels in mixed-precision (FP16) for both BPTT and CD-RGE training. BPTT is trained using AdamW (weight decay 0.1, $\beta=(0.99, 0.999)$), as recommended by \citet{graves2014neural}, and batch size 1024. For CD-RGE, we do not use weight decay. FlashRNN does not support DNCs yet, so we use a custom PyTorch implementation. We overfit a fixed batch of length 100 that is random using a batch size of 1 and FP32 precision and take the average of 5 runs. No weight decay, momentum, or learning rate schedule is used for either solver. For all Transduction experiments, we train on input sequences of length 1 to 10 and validate on sequences of length 11 to 60 where every sample is random length and randomly sampled characters. For language modeling, all models are trained with sequence length 10 via teacher forcing, and batch size 1024.

\section{Conclusion}

We present CD-RGE as a more scalable, memory-efficient approach for training large RNNs on long sequences. RGE sidesteps the limitations of BPTT, avoids retention of activations, while enabling strong parallelism. There are many exciting directions to take this work. As transformers take on more of human's mental workload, their compute, memory, and energy demands scale unsustainably. The environmental footprint of our AI use can be greatly reduced by using more efficient inference algorithms. RNNs may be able to reduce this impact and Zero-Order could be the way we train them. This work is a small step towards training large, novel RNN architectures to compete directly with transformers at scale, without the need of differentiability, continuity, and activation retention requirements.

\section{Acknowledgments}

We thank Dr. John Duchi for helpful discussions and feedback.

\bibliographystyle{plainnat}
\bibliography{references}  

\appendix

\section{Performance Characteristics of CD-RGE, BPTT, and transformers}
\begin{figure}[H]
\centering

\begin{subfigure}[b]{0.48\textwidth}
    \centering
    \includegraphics[width=\textwidth]{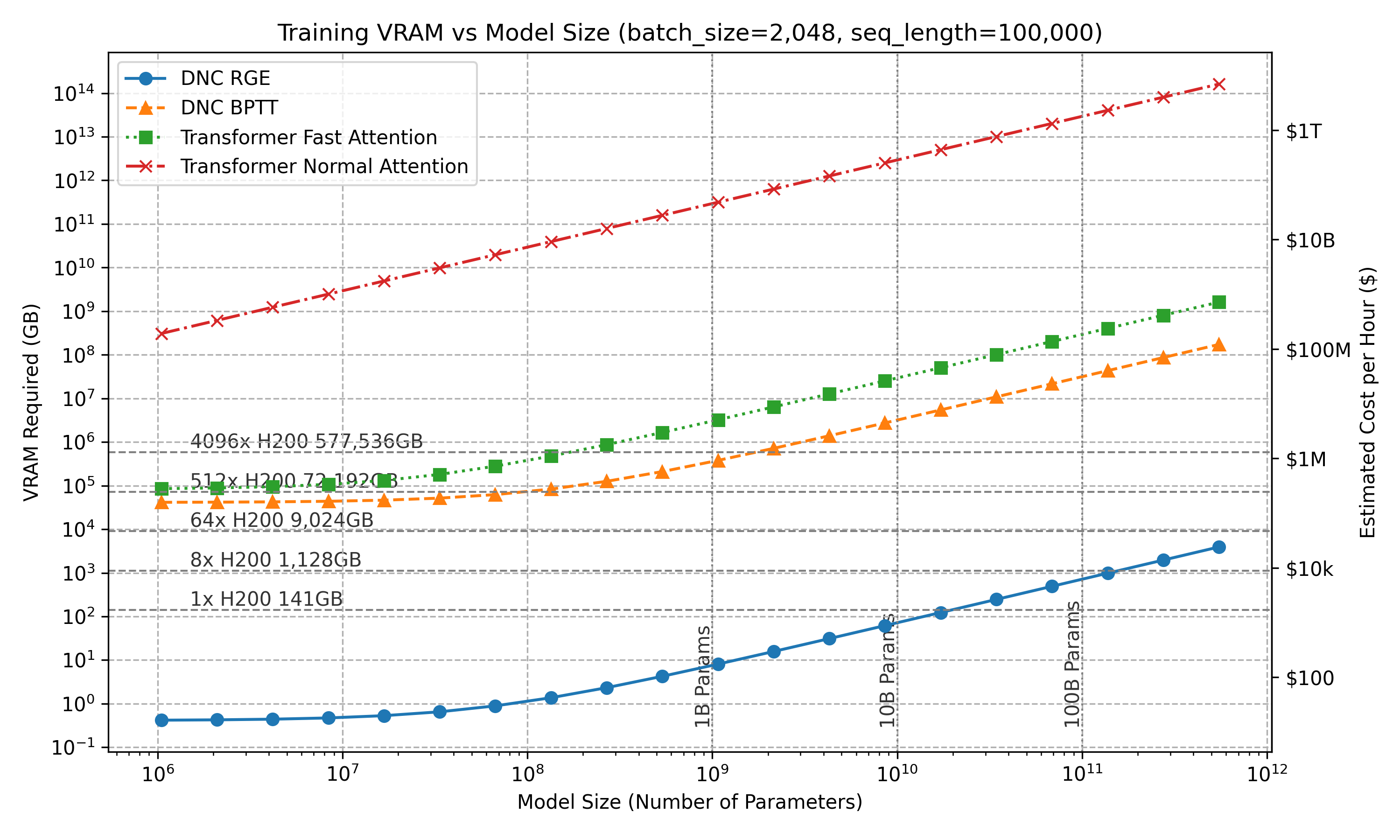}
    \caption{Training VRAM vs model size}
\end{subfigure}
\hfill
\begin{subfigure}[b]{0.48\textwidth}
    \centering
    \includegraphics[width=\textwidth]{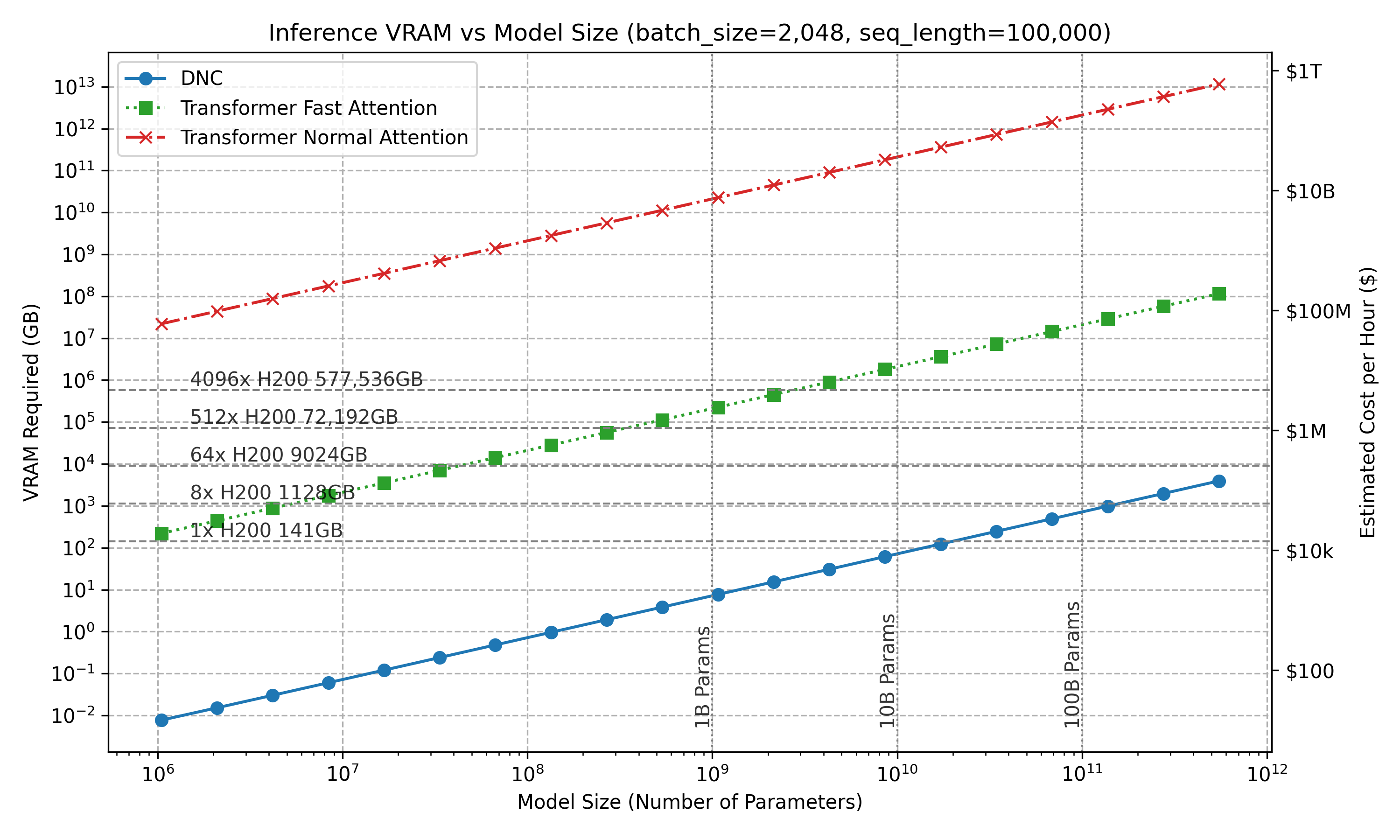}
    \caption{Inference VRAM vs model size}
\end{subfigure}

\vspace{1.5em}

\begin{subfigure}[b]{0.48\textwidth}
    \centering
    \includegraphics[width=\textwidth]{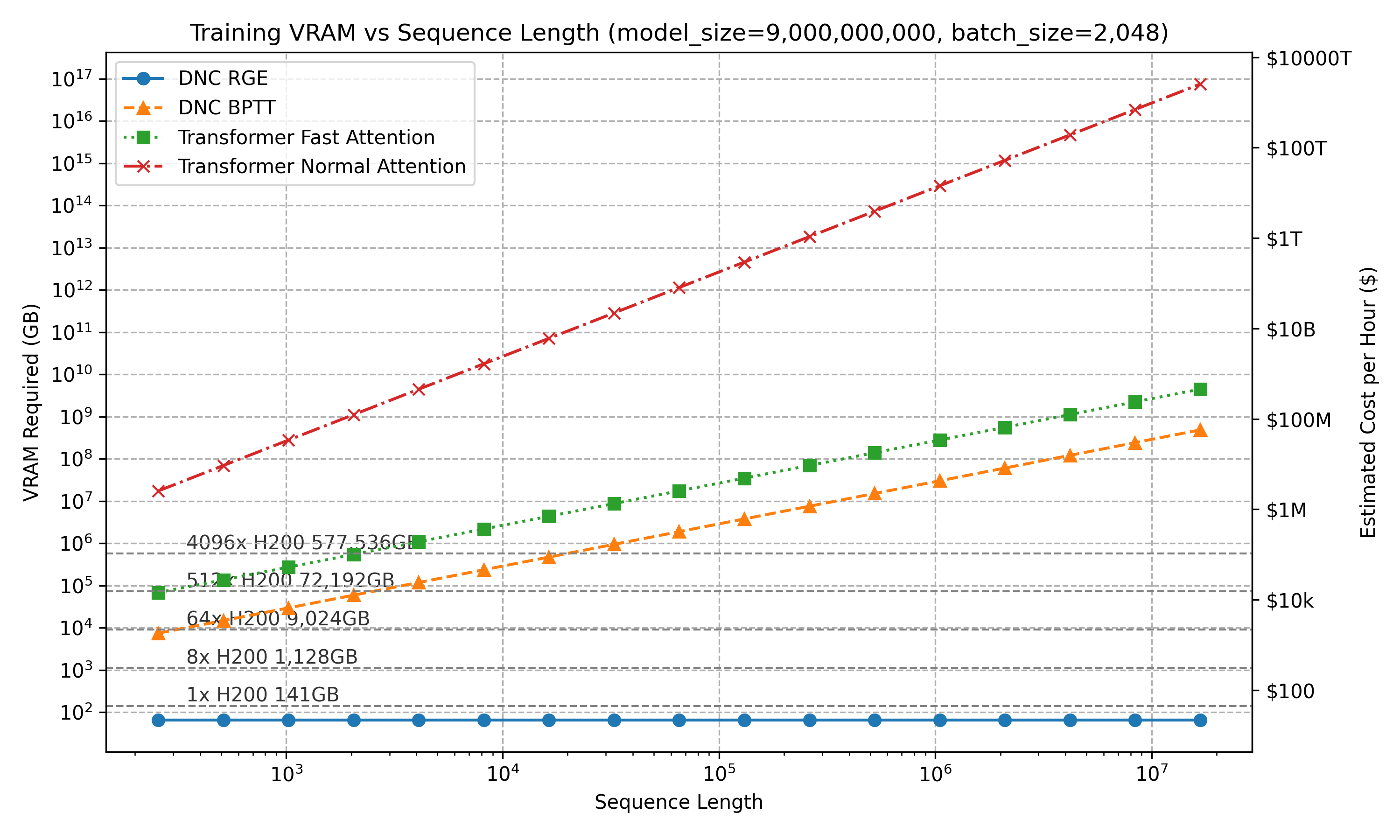}
    \caption{Training VRAM vs sequence length}
\end{subfigure}
\hfill
\begin{subfigure}[b]{0.48\textwidth}
    \centering
    \includegraphics[width=\textwidth]{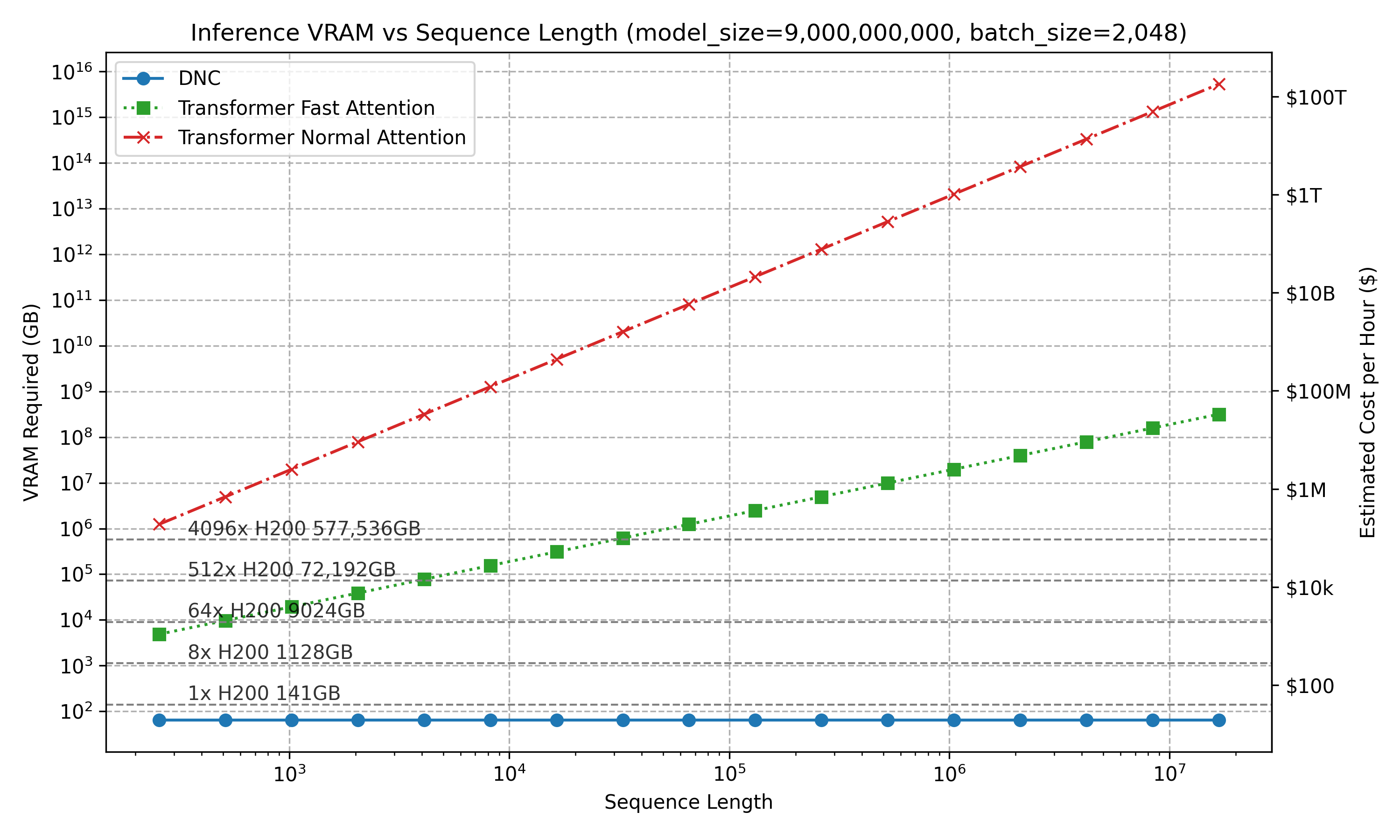}
    \caption{Inference VRAM vs sequence length}
\end{subfigure}

\vspace{1.5em}

\begin{subfigure}[b]{0.48\textwidth}
    \centering
    \includegraphics[width=\textwidth]{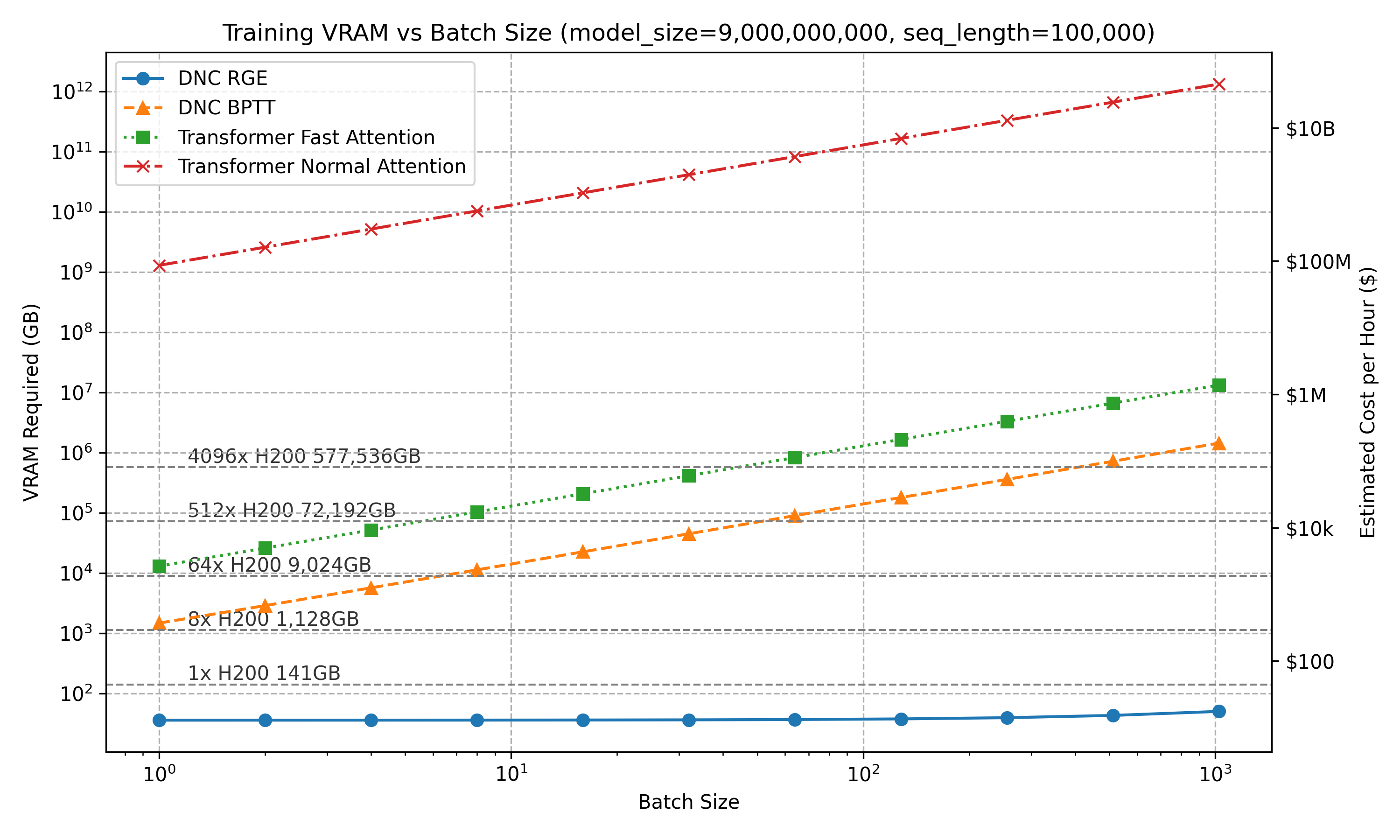}
    \caption{Training VRAM vs batch size}
\end{subfigure}
\hfill
\begin{subfigure}[b]{0.48\textwidth}
    \centering
    \includegraphics[width=\textwidth]{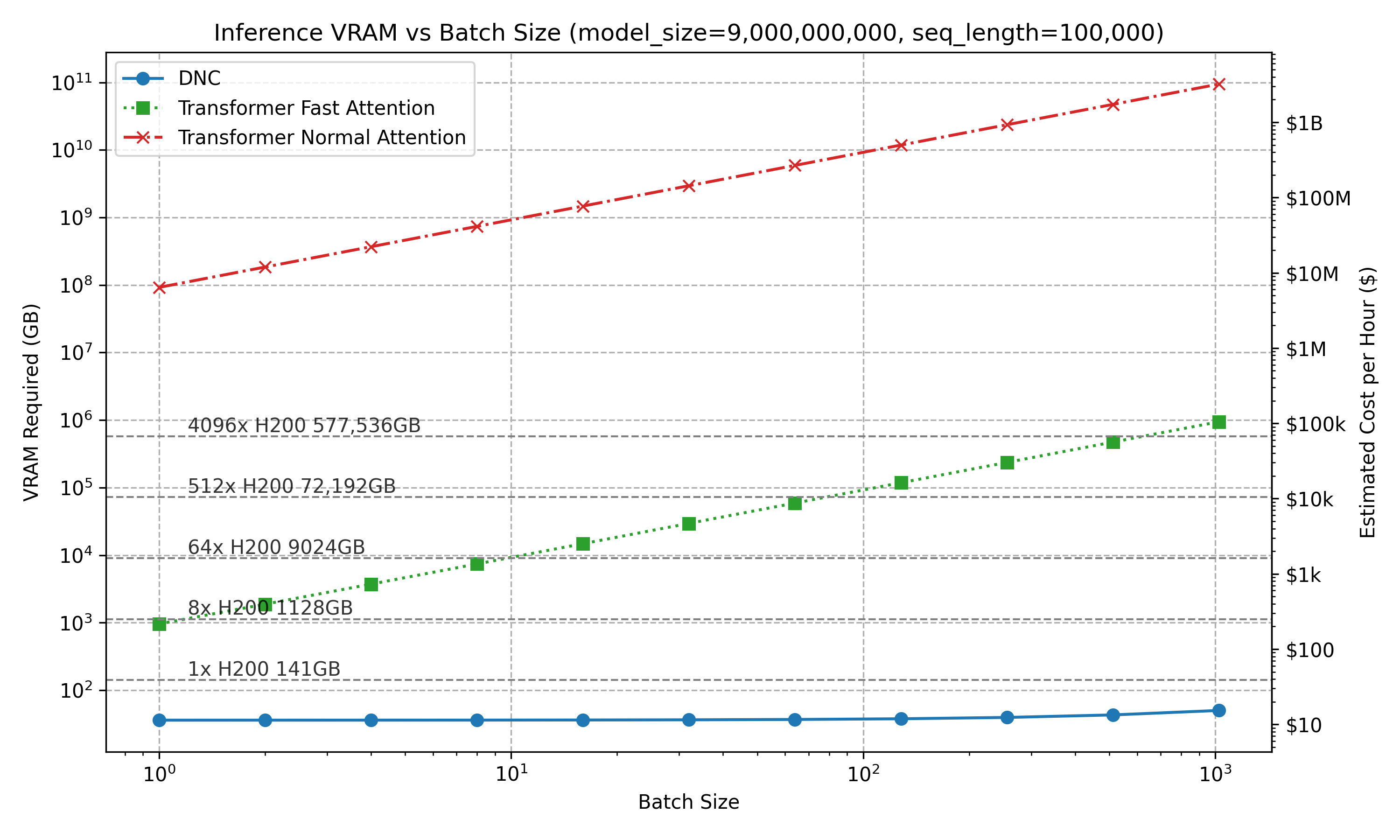}
    \caption{Inference VRAM vs batch size}
\end{subfigure}

\caption{
Comparison of VRAM requirements for training (left) and inference (right) with CD-RGE. Each row varies a key factor — model size, sequence length, or batch size — while keeping others fixed.
}
\label{fig:vram_fdras_grid}
\end{figure}

\newpage

\section{Example Smoothing Visualization: Ackley's Function Convolved with Rademacher Kernel Scaled $\varepsilon$; sweeping over $\varepsilon$=[0.1 to 4.5]}
\label{appendix:ackleys_visualization}
\begin{figure}[h]
\centering
\includegraphics[width=0.6\textwidth]{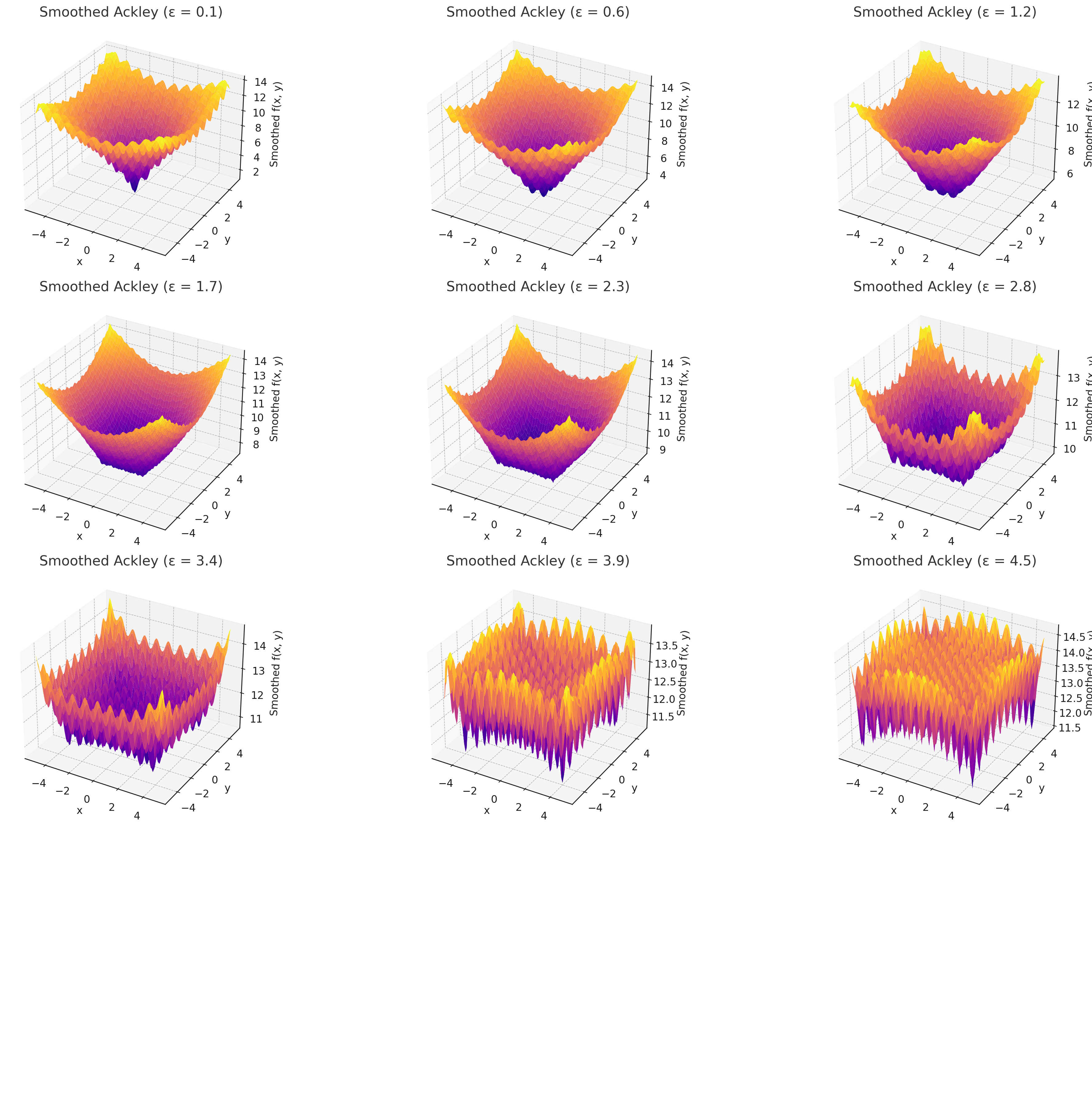} 
\caption{
Ackleys function convolved with a Rademacher distribution scaled by $\varepsilon$. As we increase $\varepsilon$, the function smooths up until a point when it becomes non-smooth again. Finding the right region is critical. For Ackleys it is around $\varepsilon$ = [1 to 1.7].
}
\label{fig:rge_shell}
\end{figure}

\section{Proof FD-RGE with antithetic probes is equivalent to CD-RGE}
\label{appendix:fdas_equals_cd}
Let $L:\mathbb{R}^{d}\!\to\!\mathbb{R}$ be differentiable, 
$\Theta\in\mathbb{R}^{d}$ be a base point, $\epsilon>0$ a stepsize,
and $p\in\mathbb{R}^{d}$ a probe direction.
Define the single–sided (forward) random-gradient estimator
\[
  \tilde g_{\mathrm{FD}}(p)
  =\frac{L(\Theta+\epsilon p)-L(\Theta)}{\epsilon}\,p,
\]
the central-difference estimator
\[
  \tilde g_{\mathrm{CD}}(p)
  =\frac{L(\Theta+\epsilon p)-L(\Theta-\epsilon p)}
         {2\,\epsilon}\,p,
\]
and the \emph{antithetic forward} estimator obtained by evaluating
both $p$ and $-p$ but still referencing the clean loss $L(\Theta)$:
\[
  \tilde g_{\mathrm{FD\!-\!AS}}(p)
  =\frac12\!\left[
      \frac{L(\Theta+\epsilon p)-L(\Theta)}{\epsilon}\,p
      +\frac{L(\Theta-\epsilon p)-L(\Theta)}{\epsilon}\,(-p)
    \right].
\]
Then for every $p$,
\[
\boxed{\;
      \tilde g_{\mathrm{FD\!-\!AS}}(p)
      \;=\;
      \tilde g_{\mathrm{CD}}(p)
      \;}
\]
and hence the two schemes have identical bias and variance.

Expand the average of the two forward differences:
\[
\begin{aligned}
  \tilde g_{\mathrm{FD\!-\!AS}}(p)
  &=\frac{1}{2\epsilon}
     \bigl[(L(\Theta+\epsilon p)-L(\Theta))\,p
           -(L(\Theta-\epsilon p)-L(\Theta))\,p\bigr] \\[4pt]
  &=\frac{p}{2\epsilon}\,
     \bigl[L(\Theta+\epsilon p)-L(\Theta-\epsilon p)\bigr] \\[4pt]
  &=\tilde g_{\mathrm{CD}}(p).
\end{aligned}
\]

\section{Visualization of probes on Rademacher Sphere and RGE selected direction}
\label{appendix:sphereplot}

\begin{figure}[H]
\centering
\includegraphics[width=0.6\textwidth]{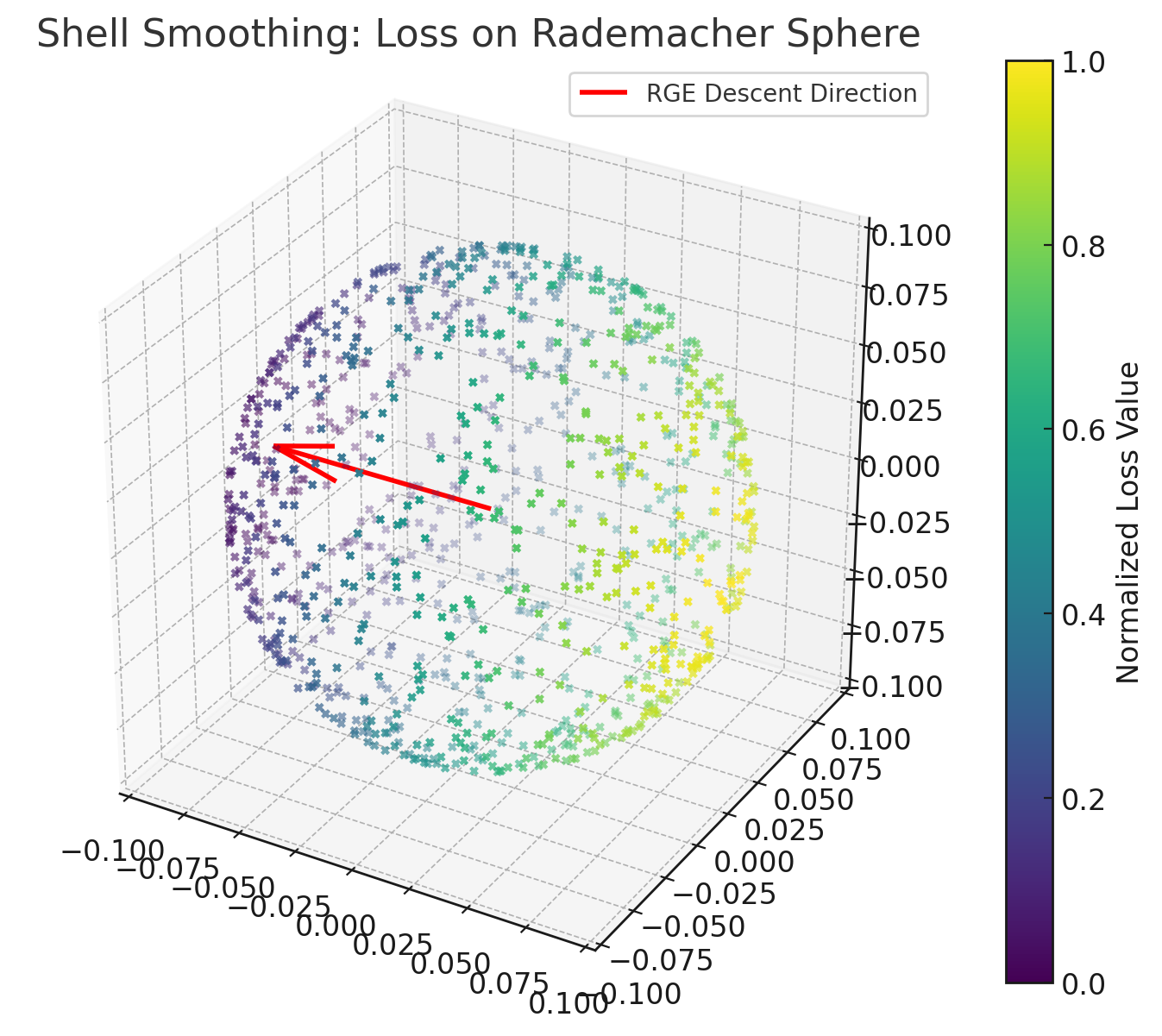} 
\caption{
Visualization of \textbf{shell smoothing in RGE} using Rademacher-like perturbations \( \epsilon p_i \) on a 3D sphere with radius \( \epsilon \sqrt{d} \). Each point represents a directional probe \( \epsilon p_i \) used to estimate the gradient via finite differences. The color denotes the loss value \( L(\theta + \epsilon p_i) \), with darker regions indicating lower loss. The red arrow shows the estimated descent direction computed via RGE as per \eqref{eq:fde}.
RGE effectively integrates directional loss differences over the shell, yielding a smooth, low-variance estimate of the gradient that consistently points toward lower-loss regions, even in noisy or non-smooth settings.
}
\label{fig:rge_shell}
\end{figure}

\end{document}